# Cross-Dataset Generalization in Deep Learning

Xuyu Zhang[1,2], Haofan Huang[3], Dawei Zhang[2], Songlin Zhuang[2], Shensheng Han[1,5], Puxiang Lai[3,4, *], and Honglin Liu[1,5, *]

[1]Shanghai Institute of Optics and Fine Mechanics, Chinese Academy of Sciences, Shanghai 201800, China

[2]School of Optical-Electrical and Computer Engineering, University of Shanghai for Science and Technology, Shanghai 200093, China

[3]Department of Biomedical Engineering, The Hong Kong Polytechnic University, Hong Kong SAR, China

[4]Photonics Research Institute, The Hong Kong Polytechnic University, Hong Kong SAR, China

[5]Center of Materials Science and Optoelectronics Engineering, University of Chinese Academy of Science, Beijing 100049, China

*puxiang.lai@polyu.edu.hk, and hlliu4@hotmail.com

**Abstract**

Deep learning has been extensively used in various fields, such as phase imaging, 3D imaging reconstruction, phase unwrapping, and laser speckle reduction, particularly for complex problems that lack analytic models. Its data-driven nature allows for implicit construction of mathematical relationships within the network through training with abundant data. However, a critical challenge in practical applications is the generalization issue, where a network trained on one dataset struggles to recognize an unknown target from a different dataset. In this study, we investigate imaging through scattering media and discover that the mathematical relationship learned by the network is an approximation dependent on the training dataset, rather than the true mapping relationship of the model. We demonstrate that enhancing the diversity of the training dataset can improve this approximation, thereby achieving generalization across different datasets, as the mapping relationship of a linear physical model is independent of inputs. This study elucidates the nature of generalization across different datasets and provides insights into the design of training datasets to ultimately address the generalization issue in various deep learning-based applications.

**Introduction**

The study of imaging through scattering media is a challenging and cutting-edge field. Scattering media are ubiquitous in everyday life, such as rough surfaces, clouds, fog, dust, water, and biological tissues. Image reconstruction through these media is particularly important in areas such as transportation, military, and biomedicine. Over the past decades, researchers have developed various methods, such as optical phase conjugation [1-3], wavefront shaping [4-9], scattering matrix measurement [11-12], and speckle correlation [13-15], to see through scattering media. In recent years, with the rapid development of artificial intelligence (AI), deep learning has emerged as a simple, efficient, and powerful tool for information extraction in this field [16-19]. Beyond this application, deep learning has been widely applied across various fields of natural science research [20-23], offering alternative solutions to complex problems. Unlike the aforementioned physics-model-based techniques, deep learning is a data-driven approach to solve inverse problems, training an implicit inversion model in the form of a neural network with a large number of paired data. The learning-based method often outperforms traditional techniques and is considered a powerful technique for addressing highly ill-posed problems in computational imaging. That said, existing learning frameworks generally rely on supervised learning strategies, which requires one-to-one input and output images. Unfortunately, in practice, obtaining data of sufficient scales and varieties is challenging. When a network is trained on a limited set, it often fails to recover images from new test data due to a lack of generalization, a ubiquitous challenge in AI [23].

Researchers have been striving to enhance the generalization capabilities of deep learning [24-27]. Although recent progress has been made and is encouraging, many strategies to improve generalization focus on capturing diverse image datasets, improving model architectures, and refining training methods. This has led to increasingly complex systems and architectures. In contrast to these strategies, we investigate the generalization issue from a physical perspective and uncover the underlying cause for the lack of generalization between different datasets.

Using a simple model of imaging through scattering media, we systematically investigated the factors affecting the generalization between datasets and proposed corresponding solutions. Previous studies have found that a network trained on a complex image dataset (e.g., face images) can predict both complex test images (face images) and simple test images (handwritten digital images), but not vice versa. Li et al. explained this phenomenon by suggesting that feature priors between different datasets affect the network's generalizability, making it impossible to identify complex features of faces through the low-order feature mapping relationship learned from handwritten digital images [28]. It is important to note that, unlike human visual object recognition, objective systems such as imaging do not distinguish target features and generally do not exhibit superior performance for one target type over another. In this context, the network should focus learning the underlying mathematical mapping of the physical system [29-31], rather than the image features of different targets in the training dataset. Therefore, attributing the network's inability to effectively predict new, unseen targets to the presence of new image features is debatable. Our study found that the mapping relationship obtained by the network after training is an approximation of the actual system's mapping. The spatial distribution and intensity distribution of effective pixels in the training images determine the gap between the network's simulated mapping and the real mapping of the system. The larger the gap, the worse the generalization. This indicates that by designing an appropriate training dataset, we can improve the network's generalization ability and lay the foundation for practical applications. This study not only addresses the problem of generalization between different datasets but also enhances the interpretability of the network to a certain extent.

**Methods**

For an imaging system, the relationship between input and output can be expressed as
$$Y = TX, \tag{1}$$
where $Y$ is the output of the system, $X$ is the input, such as the light intensity distribution of an object, and $T$ is the system's transmission matrix or optical transfer function. Of course, it is a broad representation of many systems. Strictly speaking, $T$ in actual systems may involve nonlinear effects or require complex mathematical models for accurate calculations. In a system of imaging through scattering media, $X$ represents the target to be observed, $T$ is the scattering transmission matrix, and $Y$ denotes the pattern recorded by the camera, which is a speckle pattern under coherent illumination.

The study of the generalization of a neural network usually deals with two unknowns: $T$ and $X$. The generalization for different $T$s is influenced by the medium type, thickness, spatial position, and experimental setup, and it is not discussed here. For input $X$, no matter it is simple letters, digit images, more complex faces, flora and fauna images, or various objects in real life, it can be represented by two elements: the morphology of the object and the grayscale value of each pixel (for simplicity, factors such as color are excluded),
$$X = X\{q_{morphology}(x_1, x_2, x_3, \ldots), q_{grayscale\ value}(x_1, x_2, x_3, \ldots)\}, \tag{2}$$

where $x_i$, $i = 1, 2, 3 ...$ represents different pixels, $q_{morphology}$ denotes various structures formed by these pixels, and $q_{grayscale\ value}$ are their grayscale values. Changing either $q_{morphology}$ or $q_{grayscale\ value}$ can adjust the complexity of the target.

According to Eq. (1), the process of predicting the target by the network can be expressed as
$$X = T^{-1}Y, \tag{3}$$
where $T^{-1}$ is the inverse of the system's transmission matrix, and $T^{-1}T = 1$. The training process of the network is to learn the mapping relationship between the speckle pattern and the target, $T^{-1}$, from a large amount of training data, thereby predicting $X$ from $Y$. For convenience, let the mapping relationship learned by the network be denoted as $M$. If $M = T^{-1}$, then under the same conditions, the network should be able to predict any target used for testing. However, if $M$ is only an approximation of $T^{-1}$, and different training data correspond to different $M$s, there may be lack of generalization between different datasets.

The experimental setup is shown in Fig. 1a. The output from a solid-state laser (MGL-III-532-200mW, Changchun New Industries Optoelectronics Tech) was expanded and then illuminates the target on a digital micromirror device (DMD, V-7001 VIS, ViALUX). Light reflected from the target was scattered by a ground glass diffuser (DG10-120-MD, Thorlabs), and the resultant speckle pattern was recorded by a CCD camera (Ace acA2440-75um, Basler). The wavelength of light source $\lambda = 532\ nm$，the object distance $u = 16\ cm$，and the image distance $v = 10\ cm$. The DMD array size is 1024×768 with a pixel size of 13.7 μm, and the CCD sensor array is 2448×2048 with a pixel size of 3.45 μm. The aperture diameter of the iris is 5 mm. By changing the target pattern displayed on the DMD and recording the corresponding speckle distribution, we can obtain a set of training data consisting of the target and the corresponding speckle pattern. The targets were selected from the LFW (Labeled Faces in the Wild) face dataset or the MNIST (Modified National Institute of Standards and Technology) handwritten digits dataset, and the test data are also chosen from these two datasets (Fig. 1b). Note that both datasets are open access and could be used with permission for research and academic dissemination [please refer to https://vis-www.cs.umass.edu/lfw/, and https://yann.lecun.com/exdb/mnist/].

It is assumed that the mapping relationships obtained by training with the face and MNIST datasets are $M_F$ and $M_N$ respectively, and $M_F \neq M_N$. To further explore the relationship between $M_F$ and $M_N$, we designed five sets of experiments to study the mapping relationships under different conditions (Fig. 1c): Case 1, train the network with face images, then test the network with face images and handwritten digit images, respectively; Case 2, train the network with handwritten digit images, then test the network with face images and handwritten digit images, respectively; Case 3, train the network with enlarged handwritten digit images, then test the network with face images and the original handwritten digit images, respectively; Case 4, train the network with enlarged handwritten digit images with added intensity fluctuations, then test the network with face images and the original handwritten digit images, respectively; Case 5, train the network with face images, then test the network with face images and handwritten digit images at original and translated positions.

To visually demonstrate the range of illuminated pixels on the object plane during training and the intensity of the training for these illuminated pixels, we used two different methods to sequentially superimpose the target images, as shown in Fig. 1b. For the MNIST images, the intensity-superimposed and saturated clipping images have black edges, indicating that pixels in these areas remained zero throughout the training process and were not effectively trained. In contrast, the normalized intensity-superimposed images show significantly higher intensity around the center compared to the edges, suggesting that the pixels near the center were more thoroughly trained. Conversely, the intensity-superimposed and saturated clipping images of the face images are completely white, and the normalized intensity-superimposed images exhibit a relatively uniform intensity distribution, indicating that all pixels were adequately trained. (See Section A in Supplementary Information (SI) for more details on the pixel distribution differences between the two datasets.)

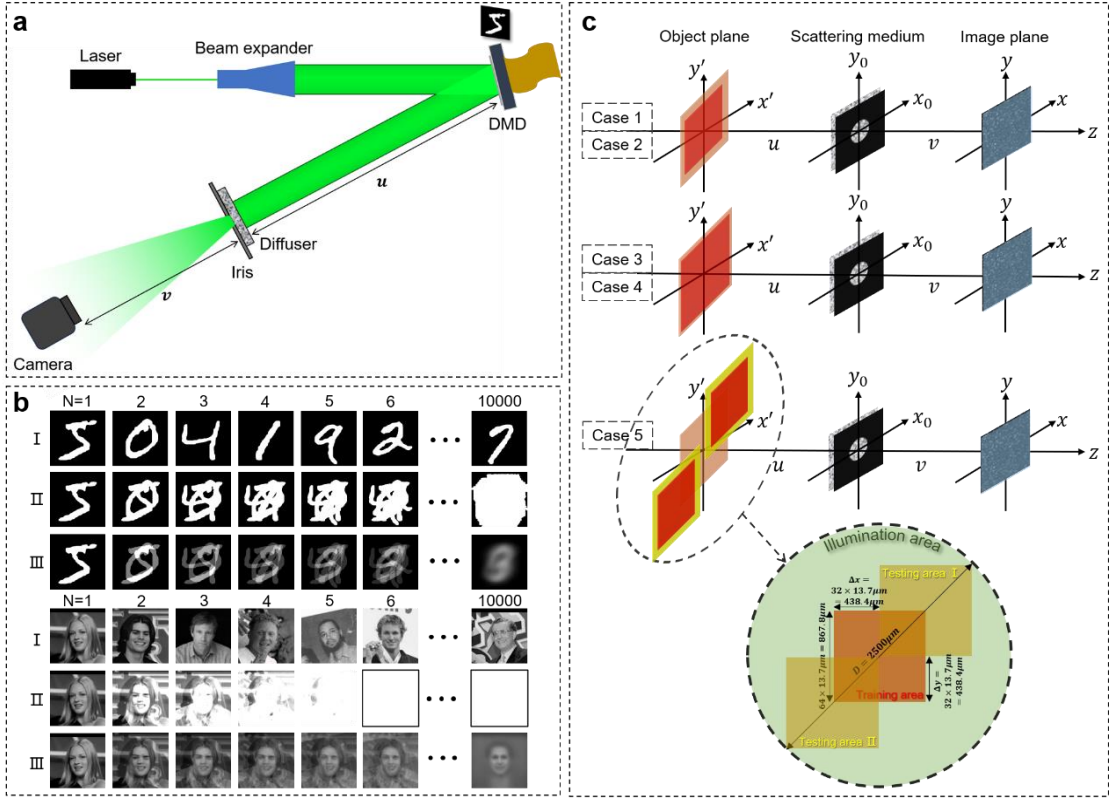

Fig.1 Experimental design. **a**, Schematic of the experimental optical setup, where $u$ is the object distance and $v$ is the image distance. The images loaded on the DMD serve as the targets. **b**, Images from the MNIST and LFW datasets are superimposed in two ways, respectively. I: Original images of handwritten digits and faces; II: Images with intensity superimposed and saturated clipping; III: Images with intensity superimposed and normalized. The black edge in the image of superimposed 10000 MNIST images indicates that these pixels haven't been activated in learning. **c**, Investigation of the mapping relationships, sequentially showing the object plane, scattering medium, and image plane in the five sets of experiments. The orange square represents the effective training region for grayscale face image targets, the slightly smaller red square denotes the effective training region for binary-handwritten-digit image targets, and the blue square represents the camera recording area. In the fifth set of experiments, the yellow squares denote shifted face-target testing areas, the smaller red squares represent the shifted binary-handwritten-digit testing areas, and the inserted green dashed circle illustrates the light spot on the DMD. All testing and training areas are within the illumination range of the light source.

The original input image arrays on the DMD were set to $64 \times 64$. In Case 3 and Case 4, the digit image arrays were enlarged to $96 \times 96$ first, and after completing the relevant operations, the central $64 \times 64$ array was cropped and used as the input image. The intensity fluctuation in Case 4 was achieved by multiplying a periodic two-dimensional distribution $P(x', y')$ expressed by

$$P(x', y') = A * [sin(k_x x' + \varphi_x) + sin(k_y y' + \varphi_y)], \qquad (4)$$

where $A \in (0,1]$, $k_x, k_y = \frac{2\pi}{16} \parallel \frac{2\pi}{24}$, $\varphi_x, \varphi_y \in [0, 2\pi)$, $\parallel$ is the logical OR operator. $(x', y')$ represents the position of a pixel within the array. To understand the variation characteristics of intensity fluctuation on training in detail, Case 4 is further divided into three scenarios, as shown in Fig. 2: a, the original digit image was enlarged and then multiplied by $P(x',y')$, with $A=1$ as a constant; b, the original digit image is enlarged and then multiplied by $P(x',y')$; c, two original digit images are superimposed, enlarged, and then multiplied by $P(x',y')$.

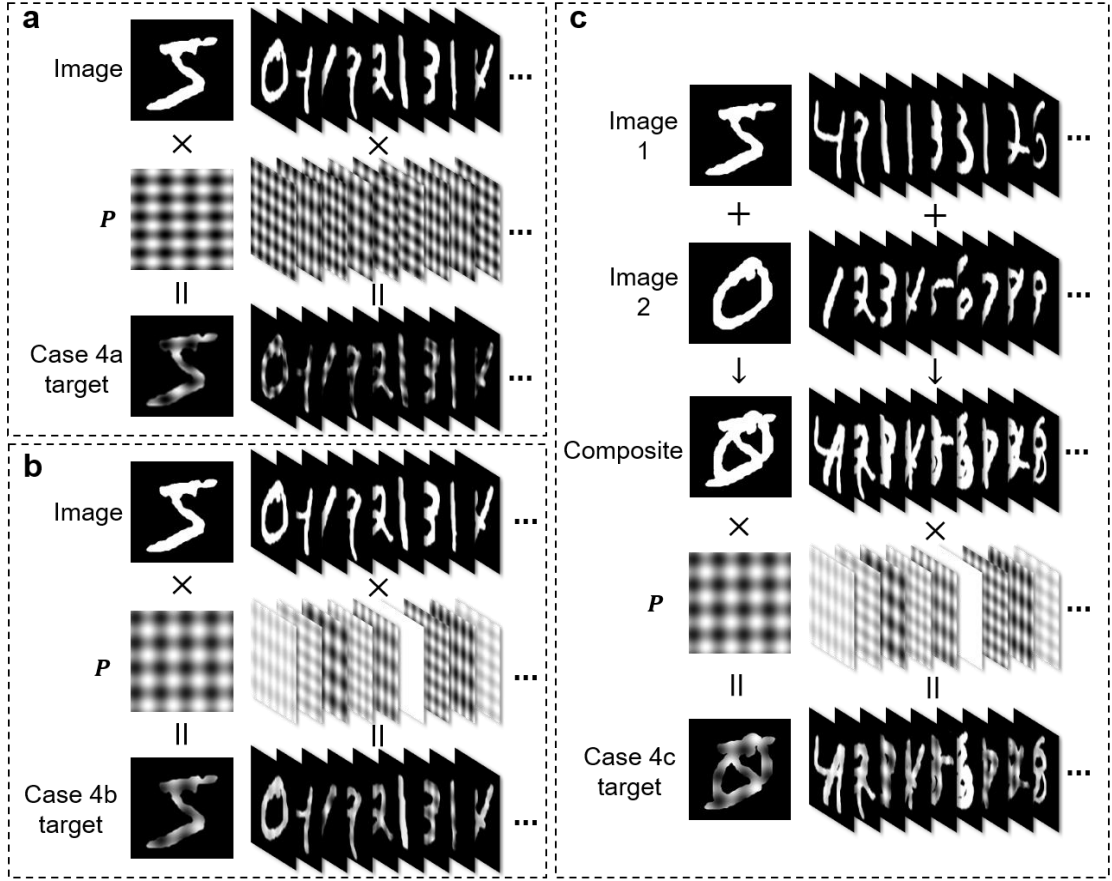

Fig.2 Illustration of the generation of inputs in Cases 4a, 4b, and 4c, respectively.

The convolutional neural network used in this study is U-Net, whose architecture can be found in Ref. [17]. The only difference is that a $7 \times 7$ convolution kernel is used instead of a $3 \times 3$ one. To improve the training efficiency, the input images are resized from a $64 \times 64$ array to a $256 \times 256$ array, and the recorded speckle patterns are downsampled from $1024 \times 1024$ to $256 \times 256$. During training, the Adam optimizer is used with an initial learning rate of 0.0001. The loss function is a weighted combination of MSE loss and Dice loss. The training is set for 50 epochs, with an early stopping strategy based on the Dice coefficient to terminate training early if necessary. Each case is trained with 9950 pairs of data. The network architecture is based on TensorFlow 2.9.0/Keras, and the experiments are run on an NVIDIA RTX 3090. In all five experiments, the network's initial state is the same. The Pearson Correlation Coefficient (PCC) is used to quantitatively evaluate the quality of the network's reconstructed images.

**Results**

As shown in Fig. 3, the network trained with the face dataset can reconstruct both face images and handwritten digits from speckle patterns (Case 1). However, the network trained with the handwritten digit dataset can only recognize the handwritten digits but not for face images (Case 2), which is consistent with the results reported in literature [28]. From the perspective of mapping relationship, $M_F$ is closer to $T^{-1}$, allowing the reconstruction of handwritten digits based on $M_F$. In contrast, $M_N$ is far away from $T^{-1}$, making it impossible to reconstruct face images based on $M_N$. In Case 1, the quality of the reconstructed face images is slightly higher than that of the digit images as told by the corresponding PCC values, whereas in Case 2, the quality of the reconstructed digit images is also slightly higher than its peers in Case 1. This indicates that the features of the training dataset are reflected in the network, that is, besides the mapping relationship, the network also learns some characteristic information of the training datasets. In Case 3, as the effective pixel area of the digit images increases, the digit images can be recognized. Although face images still cannot be effectively reconstructed due to the binary nature of the pixel values (0 and 1), the mapping relationship learned by the network is too coarse to recognize grayscale information. That said, the PCC coefficient of the reconstructed images significantly improves from its peer in Case 3. As the pixel values of the digit images become diversified, transitioning from binary to grayscale, the performance of the trained network improves further, enabling it to reconstruct face images with recognizable facial features (Case 4).

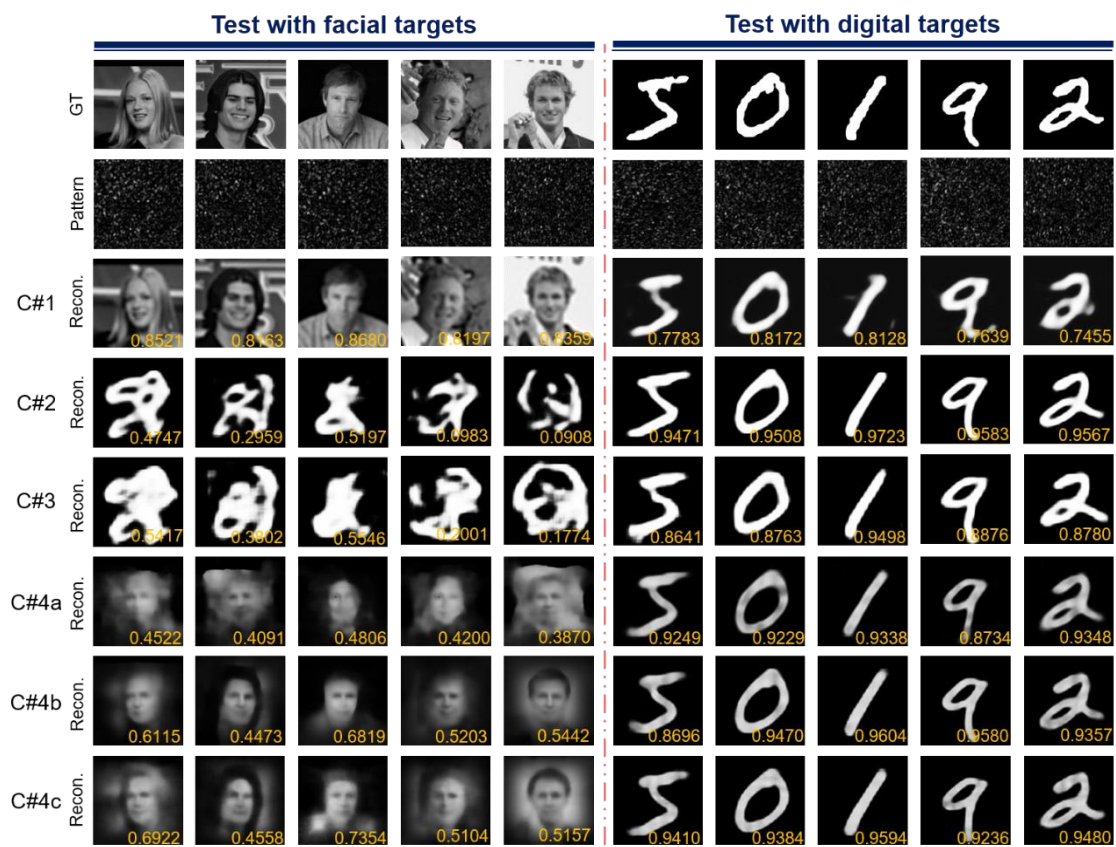

Fig.3 Reconstruction results of Cases 1-4. The first row shows the ground truth (GT) target images. The second row shows the speckle patterns for testing (Pattern). C#1-C#4c show the reconstructed results under different conditions, with PCC value with respect to the ground truth marked at the bottom right corner. Please refer to Section B in SI for more quantification analysis of reconstructed images.

Moreover, as seen when the grayscale levels are more finely tuned and exhibit a wider range of shades, the resulting reconstructed images are of better quality (Case 4a-c). On the other hand, due to more thorough training in the central region of the digit images (Fig. 1b III), the recoverable parts of the face images are also concentrated in the central region. These results demonstrate that if the effective pixels of the training dataset images fill the entire training area and the grayscale values of each pixel vary randomly, $M_N$ will increasingly approximate $T^{-1}$. Thus, a network trained with digit images can successfully predict face test data, regardless of whether the test data and training data are of the same type or have consistent features. In other words, the mapping relationship $M$ learned by the network depends on the training data; by designing the training dataset appropriately, $M$ can be as close to the ideal $T^{-1}$ as possible, reducing the network's sensitivity to the test data type and enhancing the network's generalization ability. This is also consistent with the principle that the response function of a simple scattering system, i.e., a linear time-invariant system, is independent of the system input.

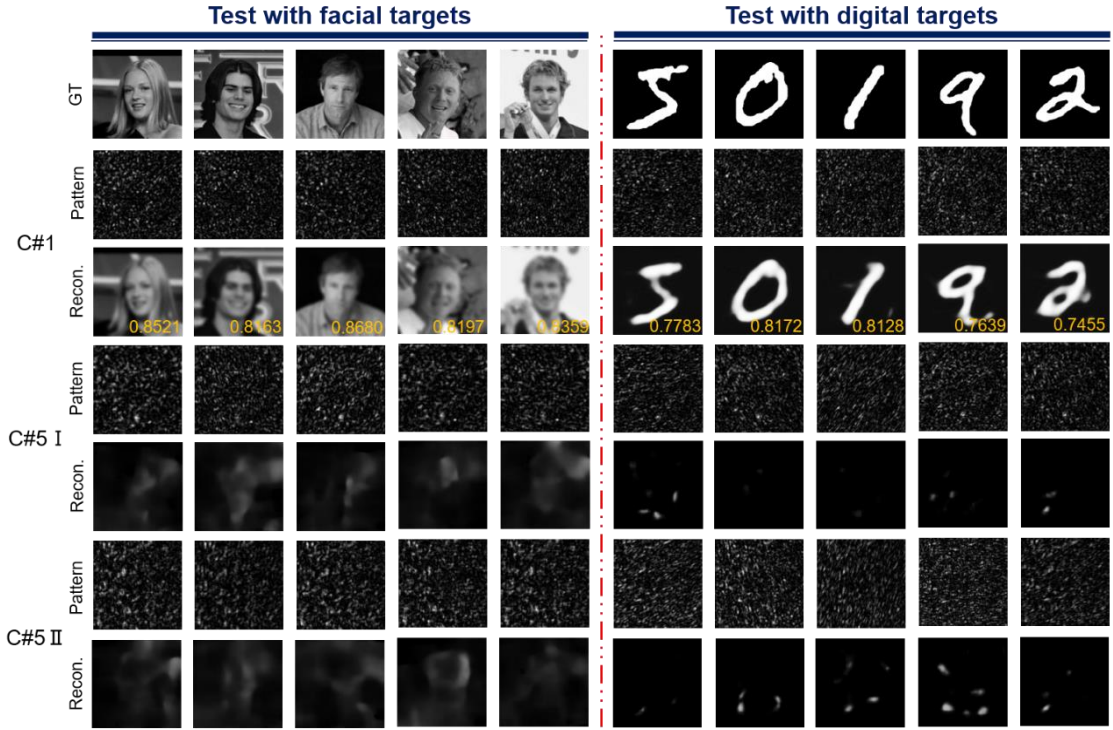

Fig.4 Reconstruction results of Case 5. The speckle patterns of original area and shifted areas I&II for testing are shown in Rows 2, 4, and 6, respectively. The PCC values of successfully reconstructed images are marked at the bottom right corner.

The results in Fig. 4 indicate that the mapping relationship learned by the network from the training data is only valid for the original regions on the object plane and the detection plane. If the test data region shifts relative to the training data, the network's predictions fail (see Section C in SI for a solution to the problem). This situation differs from using partial or down-sampled speckle patterns for image reconstruction [32, 33], where the target region and the region on the detection plane remain stationary, and the mapping relationship remains unchanged. It also should be noted that predictions based on partial or down-sampled speckle patterns only use part of the mapping relationship, although different specific parts of the mapping relationship are utilized in these two scenarios.

The position-sensitive results highlight the limitations of data-driven deep learning. The network only learns the mapping relationship within the regions covered by the training data. If the input or output regions shift, the new mapping relationship differs from the original one. The data-driven approach is inherently data-dependent. Since the learned mapping relationship is embedded within the network without explicit expression, it is challenging to infer the overall mapping relationship between the input and output planes based on the learned mapping corresponding to limited regions.

**Discussions**

Generalization is a crucial prerequisite for transitioning deep learning from laboratory settings to practical applications, and researchers have been diligently working towards this goal. One of the most commonly employed strategies to enhance network generalization is image augmentation [34-36], which involves introducing random changes to the training images, such as rotations, translations, scaling, and color adjustments. By generating similar but distinct training samples, image augmentation effectively expands the training dataset, enabling the network to learn a broader range of features and patterns. This results in better adaptability to new, unseen data. Some researchers believe that these random alterations reduce the network's dependency on specific attributes of the training data, thereby improving its generalization ability. Our research provides a deeper understanding of why image augmentation methods enhance generalization. By introducing variability in the training data, the network is forced to learn more robust and invariant features, which contribute to its improved performance on diverse test data. More importantly, our research introduces physical constraints into the field of deep learning, offering a degree of interpretability that has often been lacking. By incorporating physical principles and constraints into the training process, we can guide the network to learn more meaningful and generalizable representations. This approach not only improves the network's performance but also provides insights into the underlying mechanisms of the learned mapping relationships.

In addition to data augmentation, other methods such as physical prior guidance [37-39] and transfer learning are also employed to improve network generalization. Physical prior guidance involves incorporating domain-specific knowledge or physical laws into the training process, effectively bypassing the need to establish mapping relationships solely through data-driven learning. This method can significantly enhance the network's performance, especially in scenarios where data is scarce or expensive to obtain. However, since physical prior guidance relies heavily on predefined rules and constraints, it is not the focus of our discussion here. Transfer learning, on the other hand, is a powerful technique that leverages model parameters trained on one task to aid in learning a new, related task. By transferring knowledge from a pre-trained model, transfer learning can reduce the need for large amounts of training data and accelerate the learning process. This approach is particularly useful in situations where collecting extensive training data is impractical or time-consuming.

In could be inferred from this study that during the training process, the network not only learns the mathematical mapping relationship of the physical system but also captures some inherent features of the training images. This dual learning process explains why the reconstruction quality is often superior when the test data and training data originate from the same dataset. The features learned from the training images act as priors, influencing the reconstruction results and enhancing the network's performance on similar data.

Despite the rapid advancements in deep learning, it is essential to recognize its inherent limitations. Deep learning is fundamentally data-driven and, consequently, data-dependent. The network can only learn the mapping relationship between the target region and the detection region based on the provided training data. This learned relationship is implicit and does not extend beyond the regions covered by the training data. To infer relationships outside these trained domains, additional corresponding data must be collected and used to train the network. This data dependency underscores the importance of comprehensive and diverse training datasets to achieve robust and generalizable models.

In brief, while deep learning has shown remarkable potential, its success hinges on the quality and diversity of the training data. Techniques such as image augmentation, physical prior guidance, and transfer learning play crucial roles in enhancing network generalization and addressing the limitations of data dependency. By continuing to explore and refine these methods, we can pave the way for more robust and interpretable deep learning models that are capable of transitioning from laboratory research to practical, real-world applications.

**Conclusion**
Using an experimental system for imaging through scattering media, we investigated the issue of the lack of generalization in deep learning methods when applied to different datasets. Our findings indicate that this lack of generalization stems from limitations in the spatial and intensity distribution of the training datasets, which cause the learned mapping relationship to deviate significantly from the actual system mapping. By improving the spatial and intensity distribution characteristics of the training data, the implicit mapping relationship learned by the network can more closely approximate the true system mapping. This enhancement in the network's generalization ability enables the prediction of different types of targets. Compared with traditional data augmentation methods, our work identifies the fundamental reason for the lack of generalization between different datasets and theoretically explains why data augmentation strategies can improve network generalization. Additionally, our study bridges the gap between deep learning and imaging through scattering media, making deep learning more interpretable and providing valuable insights for its application in various fields. The study offers theoretical guidance for the design of training datasets and paves a new path for research in this area. It also has the potential to significantly boost the applications of AI in real-world scenarios. By enhancing the generalization capabilities of deep learning models, our work contributes to the broader adoption and effectiveness of AI technologies across diverse practical applications.

**Funding**
The work was supported by National Natural Science Foundation of China (NSFC) (81930048), Guangdong Science and Technology Commission (2019BT02X105), Hong Kong Research Grant Council (15217721, 15125724, C7074-21GF), Shenzhen Science and Technology Innovation Commission (JCYJ20220818100202005), and The Hong Kong Polytechnic University (P0045680, P0043485, P0045762, P0049101).

**Acknowledgements**
H. L. conceived the idea and designed the study. X. Z. implemented the experiment and simulation. H. L. X. Z., H. H., and P. L. analyzed the data and wrote the manuscript. All contribute to revising the manuscript.

**Competing Interests**
The authors declare no conflict of interests.

# Supplementary Information for "Cross-Dataset Generalization in Deep Learning"

Xuyu Zhang[1,2], Haofan Huang[3], Dawei Zhang[2], Songlin Zhuang[2], Shensheng Han[1,5], Puxiang Lai[3,4, *], and Honglin Liu[1,5, *]

[1]*Shanghai Institute of Optics and Fine Mechanics, Chinese Academy of Sciences, Shanghai 201800, China*

[2]*School of Optical-Electrical and Computer Engineering, University of Shanghai for Science and Technology, Shanghai 200093, China*

[3]*Department of Biomedical Engineering, The Hong Kong Polytechnic University, Hong Kong SAR, China*

[4]*Photonics Research Institute, The Hong Kong Polytechnic University, Hong Kong SAR, China*

[5]*Center of Materials Science and Optoelectronics Engineering, University of Chinese Academy of Science, Beijing 100049, China*

*\*puxiang.lai@polyu.edu.hk, and hlliu4@hotmail.com*

### A.  Comparison of the characteristics of LFW and MNIST datasets

The characteristics of LFW and MNIST datasets used in the study are analyzed based on histograms of pixel values. The images are 256×256 arrays in 8-bit depth greyscale. Fig.S1a shows an example of face image, where 10 pixels are randomly selected and marked. Fig.S1b shows the normalized histograms of pixel value of Points 1-10 across 9950 face images. It can be observed that the histograms vary among different pixels, with pixels on the face having higher values. Fig.S1c is the arithmetic average of the ten histograms, which is continuous with a nearly uniform distribution.

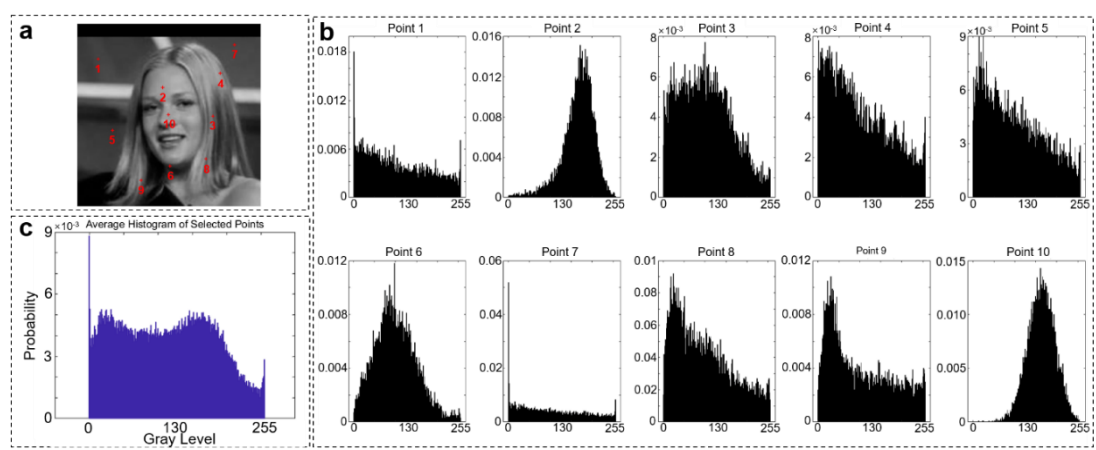

Fig.S1 Histograms of pixel value of face images in LFW dataset. **a**, an example of face image from LFW dataset with 10 pixels marked. **b**, normalized histograms of Points 1-10, respectively. **c**, arithmetic average of the statistical histograms for the 10 points in b.

The histograms of original and intensity modified MNIST dataset are shown in Fig.S2 and Fig.S3, respectively. The same 10 pixels are selected. There are two spikes at 0 and 255 in the histograms of the 10 pixels except Points 1&7 (see Fig.S2), who have only one spike at 0 since it is always dark at the edge of digit images. Their average in Fig.S2c also has two peaks with higher peak at 0. In Fig.S3, the 0 value will not be changed by intensity modification, hence, the spike at 0 remains almost the same. The spike at 255 is diffused over 0-255, with a continuous and relatively uniform distribution. It can be seen that, apart from the 0 values, the remaining histograms are continuous, showing a trend towards the nearly uniform distribution, although there are still differences overall.

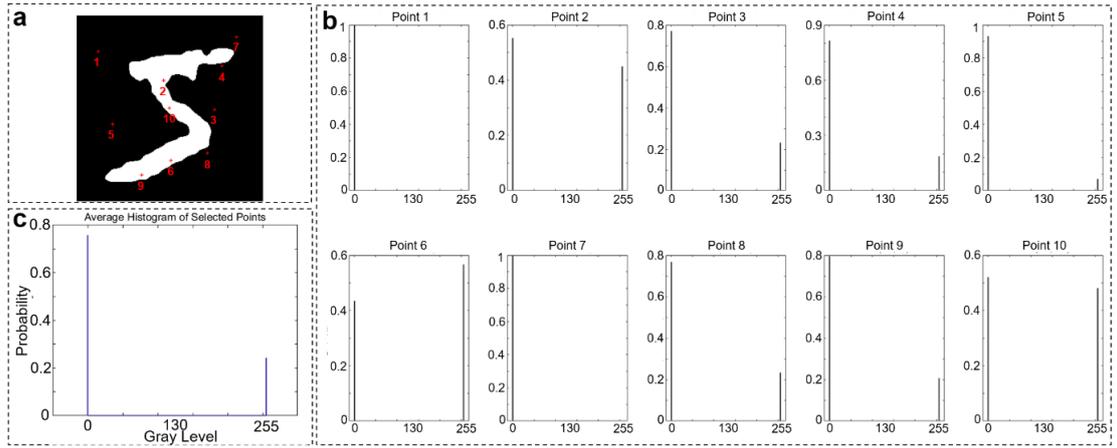

Fig.S2 Histograms of pixel value of digit images in MNIST dataset. **a**, an example of digit image from MNIST dataset with 10 pixels marked. **b**, normalized histograms of Points 1-10, respectively. **c**, arithmetic average of the statistical histograms for the 10 points in b.

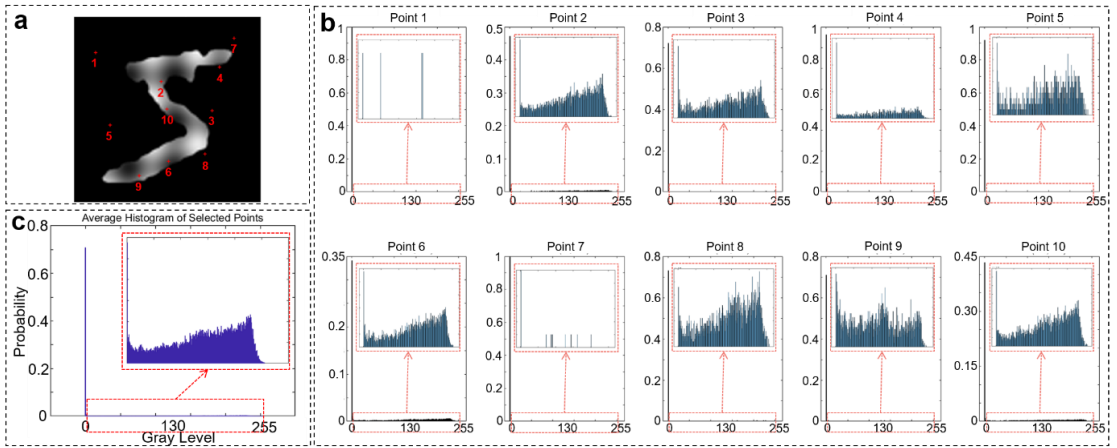

Fig.S3 Histograms of pixel value of intensity-modified digit images. **a**, an example of the intensity-modified digit image with 10 pixels marked. **b**, normalized histograms of Points 1-10, respectively. **c**, arithmetic average of the statistical histograms for the 10 points in b. The insert in each histogram is the amplified histogram from 1 to 255.

For the enlarged and intensity-modified digit images in Case 4, scaling from $64 \times 64$ to $96 \times 96$ corresponding to reshaping the $256 \times 256$ arrays to a $384 \times 384$ one. The positions of the 10 pixels in the central areas are denoted in Fig.S4a. The corresponding results are shown in Fig.S4. When the target is enlarged, the areas with intensity fluctuations also become larger, making the intensity variations within the region used for calculating histogram more pronounced and increasing the proportion of non-zero grayscale areas, leading to reduced height of 0 peak. By comparing the localized enlargements in Fig.S4c and Fig.S3c, it is evident that enlarging the MNIST image increases the complexity of pixel intensities within the original training region, making it more similar to the grayscale distribution of face images. Consequently, the results in Case 4 of the main text show that the enlarged images are more favorable for reconstructing face test images.

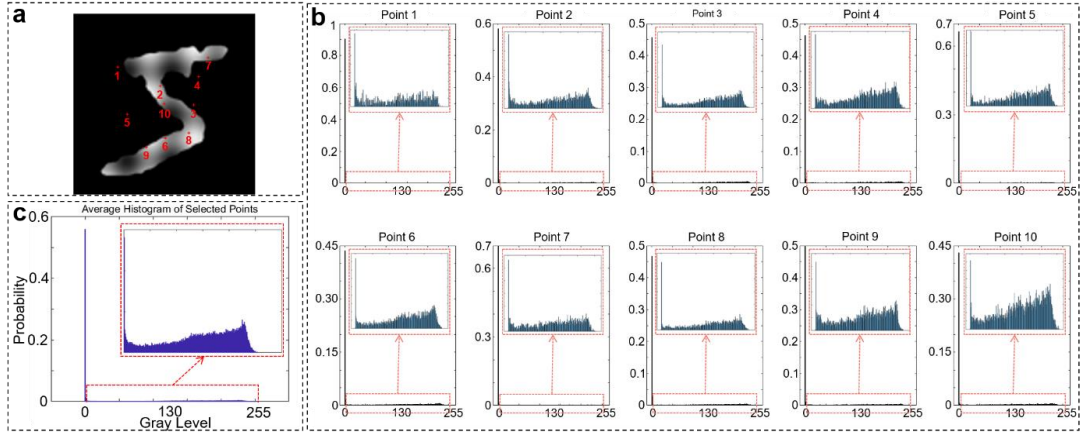

Fig.S4 Histograms of pixel value of enlarged and intensity-modified digit images. **a**, an example of the digit image with 10 pixels marked. **b**, normalized histograms of Points 1-10, respectively. **c**, arithmetic average of the statistical histograms for the 10 points in b. The insert in each histogram is the amplified histogram from 1 to 255.

**B. Comparison of the image reconstruction quality with different parameters**

Fig.3 in the main text has demonstrated that the network trained with dataset with more diversities has better generalization from cases 1 to 4, along with the corresponding PCC values. Here, we have incorporated two additional metrics commonly used to quantify the quality of reconstructed images—Structural Similarity Index (SSIM) and Cosine Similarity (CS)—to further illustrate the trend observed in our study, where the testing results progressively improve as the dataset evolves. Table 1 presents the PCC, SSIM, and Cosine Similarity values for each reconstructed image compared to its corresponding ground truth, following the same layout format of Fig.3 in the main text. Furthermore, the specific distribution of these metrics is visualized as scatter plots in Fig.S5, with the optimal values for the facial images highlighted in bold. It is evident that, whether considering SSIM or Cosine Similarity, the trend in the testing results from cases 1 to 4 is similar to that of the PCC values, albeit with slight variations in magnitude.

Table 1 PCC, SSIM, and Cosine Similarity values for each reconstructed image compared to its corresponding ground truth.

|  |  | Test with facial targets |  |  |  |  | Test with digital targets |  |  |  |  |
|---|---|---|---|---|---|---|---|---|---|---|---|
| P C C | C#1 | 0.8521 | 0.8163 | 0.8680 | 0.8197 | 0.8359 | 0.7783 | 0.8172 | 0.8128 | 0.7639 | 0.7455 |
|  | C#2 | 0.4747 | 0.2959 | 0.5197 | 0.0983 | 0.0908 | 0.9471 | 0.9508 | 0.9723 | 0.9583 | 0.9567 |
|  | C#3 | 0.5417 | 0.3802 | 0.5546 | 0.2001 | 0.1774 | 0.8641 | 0.8763 | 0.9498 | 0.8876 | 0.8780 |
|  | C#4a | 0.4522 | 0.4091 | 0.4806 | 0.2400 | 0.3870 | 0.9249 | 0.9229 | 0.9338 | 0.8734 | 0.9348 |
|  | C#4b | 0.6115 | 0.4473 | 0.6819 | 0.5203 | 0.5442 | 0.8696 | 0.9470 | 0.9604 | 0.9580 | 0.9357 |
|  | C#4c | 0.6922 | 0.4558 | 0.7354 | 0.5104 | 0.5157 | 0.9410 | 0.9384 | 0.9594 | 0.9236 | 0.9480 |
| S S I M | C#1 | 0.8367 | 0.8109 | 0.8648 | 0.8162 | 0.8391 | 0.2419 | 0.2728 | 0.3102 | 0.2582 | 0.1529 |
|  | C#2 | 0.0208 | 0.0307 | 0.1682 | 0.0259 | 0.0227 | 0.6005 | 0.6456 | 0.7551 | 0.6705 | 0.6148 |
|  | C#3 | 0.0576 | 0.0390 | 0.2106 | 0.0285 | 0.0393 | 0.4759 | 0.4658 | 0.6989 | 0.5983 | 0.5709 |
|  | C#4a | 0.1351 | 0.0623 | 0.2772 | 0.1106 | 0.1061 | 0.4796 | 0.5077 | 0.6582 | 0.4447 | 0.4122 |
|  | C#4b | 0.4051 | 0.3741 | 0.4896 | 0.3161 | 0.2533 | 0.5903 | 0.6277 | 0.7726 | 0.7365 | 0.6747 |
|  | C#4c | 0.4282 | 0.3622 | 0.4241 | 0.3427 | 0.3131 | 0.5835 | 0.6049 | 0.7683 | 0.7182 | 0.6630 |
| C S | C#1 | 0.9834 | 0.9729 | 0.9779 | 0.9735 | 0.9853 | 0.6688 | 0.6745 | 0.6724 | 0.6609 | 0.6139 |
|  | C#2 | 0.5684 | 0.3952 | 0.6374 | 0.4776 | 0.4918 | 0.7441 | 0.6891 | 0.7161 | 0.6439 | 0.7288 |
|  | C#3 | 0.6226 | 0.5397 | 0.7154 | 0.5377 | 0.6245 | 0.6279 | 0.5176 | 0.6253 | 0.5836 | 0.5920 |
|  | C#4a | 0.8023 | 0.5454 | 0.6956 | 0.6109 | 0.7236 | 0.6598 | 0.5498 | 0.6106 | 0.6017 | 0.6066 |
|  | C#4b | 0.8898 | 0.8152 | 0.8406 | 0.8072 | 0.7993 | 0.5735 | 0.5923 | 0.6478 | 0.7178 | 0.6372 |
|  | C#4c | 0.9263 | 0.8252 | 0.8879 | 0.8137 | 0.8569 | 0.6933 | 0.5386 | 0.6126 | 0.5842 | 0.7096 |

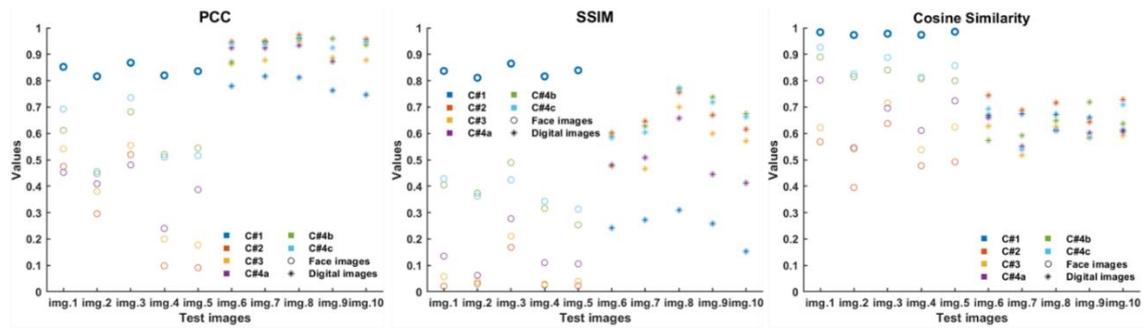

Fig.S5 From left to right, the diagram sequentially presents the PCC values, SSIM values, and Cosine Similarity for the testing results of cases 1 to case 4, as shown in Fig.3 of the main text. "img.1" to "img.10" correspond to the ten facial images and handwritten digit images used for testing in Fig.3, listed from left to right, with different colors representing different cases. Circles and asterisks are used to denote testing with facial images and handwritten digit images, respectively. The optimal values for the facial image tests are highlighted in bold.

**C. Prediction with network trained with randomly distributed digit images.**

In case 5, it has been demonstrated that regions of the object plane that were not trained by the network could not be predicted, indicating that the network had not yet learned the mapping relationship $A$ for those regions. To enhance the comprehensiveness of this conclusion, we conducted a supplementary experiment. In our experiments, the central illumination range was approximately a 128×128 array, whereas in case 5, only the central 64×64 region is used to display the targets. In the supplementary experiment, we randomly display the 64×64 targets within the 128×128 training range and collected corresponding speckle patterns for training. The schematic of the data collection strategy is depicted in Fig.S6. After training, we tested the network by randomly positioning the same digit image "5" in four different locations within the illumination range. The results, shown in Fig.S6b, include the PCC values between the reconstructions and the ground truth, highlighted in yellow in the lower right corner of each image. The test results reveal that due to the random placement of the handwritten digit targets during training, a larger portion of the object plane within the illumination range was illuminated, greatly exceeding the central 64×64 region used in case 5. As a result, the network successfully established a mapping relationship $M'$ over a broader area on object plane with the detection area, enabling accurate reconstruction of targets at various positions within the illumination range with high similarity.

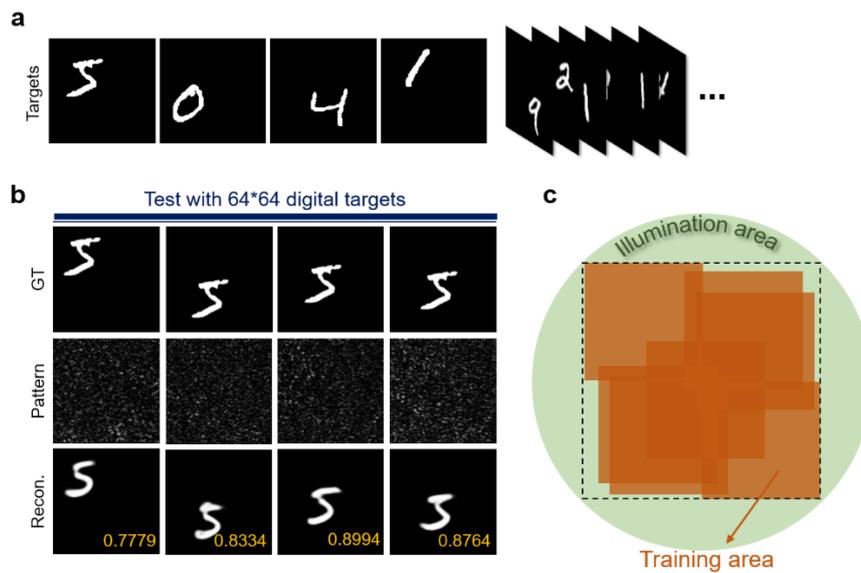

Fig.S6 Results of the supplementary experiment corresponding to case 5. **a**, a set of target images was generated by randomly placing 64×64 handwritten digit images within the 128×128 range, and corresponding speckle patterns were collected to train the network. **b**, the network testing results of digit 5 at 4 different positions, where the yellow numbers in the lower right corner of the reconstructed images are the PCC values. **c**, a schematic illustration of the target displaying area and the illumination range, where the orange training areas are the regions of training targets, and the dash square denotes the 128×128 training area.